\documentclass[11pt,a4paper]{article}
\usepackage[hyperref]{AACL-IJCNLP2020}
\usepackage{times}
\usepackage{latexsym}
\usepackage{algorithm}
\usepackage{graphicx}
\usepackage{amsmath}

\usepackage{subfig}

\aclfinalcopy % Uncomment this line for the final submission
\title{Deep Active Learning for Sequence Labeling 
\\ Based on Diversity and Uncertainty in Gradient}

\author{Yekyung Kim \\
  LG Electronics AI Lab\\
  \texttt{yekyung.kim@lge.com} \\}

\date{}

\begin{document}
\maketitle

\begin{abstract}

Recently, several studies have investigated active learning (AL) for natural language processing tasks to alleviate data dependency. However, for query selection, most of these studies mainly rely on uncertainty-based sampling, which generally does not exploit the structural information of the unlabeled data. This leads to a sampling bias in the batch active learning setting, which selects several samples at once. In this work, we demonstrate that the amount of labeled training data can be reduced using active learning when it incorporates both uncertainty and diversity in the sequence labeling task. We examined the effects of our sequence-based approach by selecting weighted diverse in the gradient embedding approach across multiple tasks, datasets, models, and consistently outperform classic uncertainty-based sampling and diversity-based sampling. 

\end{abstract}

\begin{table*}
 \centering
\begin{tabular}{c|ccccccc}
\hline
Utterance    & show & me & flight & from & Boston     & to & Denver     \\ \hline
Slot label   & O    & O  & O      & O    & B-fromloc  & O  & B-toloc    \\
Entity label & O    & O  & O      & O    & B-Location & O  & B-Location \\ \hline
\end{tabular}
\caption{An example of utterance and the annotation}
\end{table*}
\section{Introduction}
Sequence labeling is one of the commonly used techniques for solving natural language understanding (NLP) tasks such as named-entity recognition (NER) and slot filling. Furthermore, for these tasks, the state-of-the-art results are typically based on deep neural networks \cite{DBLP:conf/emnlp/KurataXZY16, DBLP:conf/interspeech/LiuL16, DBLP:conf/acl/MaH16}. However, the performance of these models is highly dependent on the availability of large amounts of annotated data. Moreover, compared with classification tasks, which require only one label for a sample, the sequence learning tasks require a series of token-level labels for an entire sequence, which makes them time-consuming and a costly annotation process. 
\\
This problem can be mitigated using active learning (AL), which achieves improved performance with fewer annotations by strategically selecting the examples to annotate \cite{DBLP:journals/jair/CohnGJ96, DBLP:conf/emnlp/SettlesC08, DBLP:conf/emnlp/SiddhantL18, DBLP:conf/iclr/ShenYLKA18}. 
There are two major strategies for active learning, namely, diversity-based sampling and uncertainty-based sampling \cite{mastersthesis, DBLP:conf/iclr/AshZK0A20}.
% 여기 이상해
Traditionally, uncertainty-based sampling is the most common pool-based AL approach. However, previous work pointed out that focusing only on the uncertainty leads to a sampling bias \cite{DBLP:journals/tcs/Dasgupta11}. It creates a pathological scenario where selected samples are highly similar to each other, which clearly indicates inefficiency. This may cause problems, especially in the case of noisy and redundant real-world datasets. Another approach is diversity-based sampling, wherein the model selects a diverse set such that it represent the input space without adding considerable redundancy \cite{DBLP:conf/iclr/SenerS18}. This approach can select samples while ensuring a maximum batch diversity. However, this approach might select points that provide little new information, thereby reducing the uncertainty of the model. Certain recent studies for classification tasks implemented an algorithm named Batch Active learning by Diverse Gradient Embeddings (BADGE). This algorithm first computes embedding for each unlabeled sample based on induced gradients, and then geometrically picks the instances from the space to ensure their diversity \cite{DBLP:conf/iclr/AshZK0A20}. 
Although it proves to be a robust improvement when performing an image classification task, its performance in sequence labeling tasks is yet unproven.
\\
In this study, we investigated some practical active learning algorithms that consider uncertainty and diversity in sequence labeling tasks over different datasets and models. Moreover, we suggested a method to expand BADGE with weighted sampling based on the sequence length to ensure cost-effective labeling. This simple modification in it has a positive implication that it tends to select cost-effective samples. 
The proposed model trades off between uncertainty and diversity by selecting diverse samples in the gradient space depending on the parameters in the final layer, for which, the currently available models focus only on uncertainty. To the best of our knowledge, our study is the first to apply diverse gradient embedding to a sequence labeling task.
We experimented with the CoNLL 2003 English, ATIS, and Facebook Multilingual Task Oriented Dataset (FMTOD). Accordingly, it was empirically demonstrated that the proposed method consistently outperformed the baseline method including Bayesian AL by disagreement (BALD), which shows state-of-the-art performance in NER task, across the datasets, tasks and model architectures.

\section{Related Work}
Several recent papers investigated AL to alleviate the data dependency of deep learning for NLP. A different query criterion based on expected gradient length (EGL) has been proposed \cite{DBLP:conf/emnlp/SettlesC08}. Furthermore, \citet{DBLP:conf/aaai/ZhangLW17} who addressed text classification, proposed selecting the examples according to the expected gradient length of the word embedding layer. \citet{DBLP:conf/emnlp/SiddhantL18, DBLP:conf/iclr/ShenYLKA18} addressed Deep Bayesian Active Learning for NER task and proved that BALD exhibits a state-of-the-arts performance. However, those works do not consider the diversity of the examples. \\
\citet{DBLP:conf/iclr/ShenYLKA18} attempted solving this problem through a hybrid AL method that performed a representativeness-based sampling weighted by uncertainty. However, it could not outperform the uncertainty-based methods. For the slot filling task, researchers proposed adversarial AL for sequence learning with an additional discriminator network \cite{DBLP:conf/ijcai/DengCSJ18} and submodularity-inspired data ranking function \cite{DBLP:journals/corr/abs-1802-00757} to select low-data regime.
Some studies have explored hybrid methods that incorporate both diversity and uncertainty in classification. \citet{DBLP:journals/corr/abs-1901-05954} takes into account both the informativeness of the examples for the model as well as their diversity in a mini-batch during classification; moreover, it considers uncertainty as a scalar value similar to works discussed earlier herein. \citet{DBLP:conf/iclr/AshZK0A20} also designed a BADGE method to incorporate both predictive uncertainty and sample
diversity. This method exhibits a robust performance in different environmental settings while performing an image classification task. In this work, we have provided an empirical evaluation of diverse gradient embeddings for NLP tasks using different models, and explored a different method to consider variable-length input.

\section{Background}
\subsection{Sequence labeling}
Slot-filling and NER are the fundamental tasks for building spoken language understanding systems. Both of these utilize sequence labeling as an approach and serve different purposes. The slot-filling task finds relevant information in a query and tags each token with the corresponding slot labels. Comparatively, NER is more generic and is likely to have fewer labels than slot-filling. A example of this alignment comparison is provided in Table 1.
In sequence labeling, the training examples map an input sequence, x, to the corresponding label sequence, y. The input of this sequence is comprised of a sequence of words, denoted as \begin{math} x = (w_1, w_2, ..., w_t )\end{math}.
Moreover, the input and label sequences have identical lengths. Therefore, there is an explicit alignment at each time step of the sentence. We used two architectures for training: the bidirectional long short- term memory (BiLSTM) tagger and the BiLSTM encoder-decoder, as described in Figure 1. 

\begin{figure}[ht]
\centering
\subfloat[Subfigure 1 list of figures text][BiLSTM tagger without neural decoder layer]{
\includegraphics[width=0.2\textwidth]{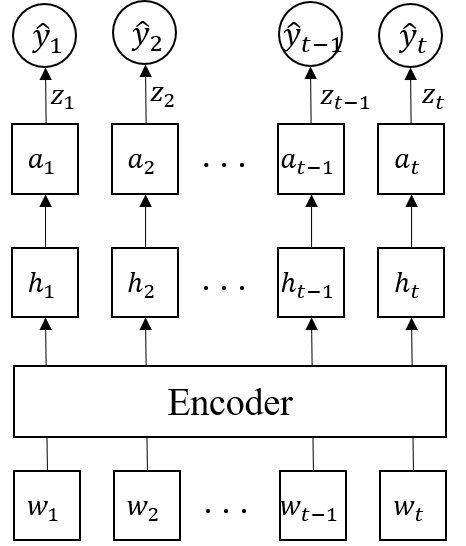}}
\qquad
\subfloat[Subfigure 2 list of figures text][BiLSTM Encoder-Decoder]{
\includegraphics[width=0.2\textwidth]{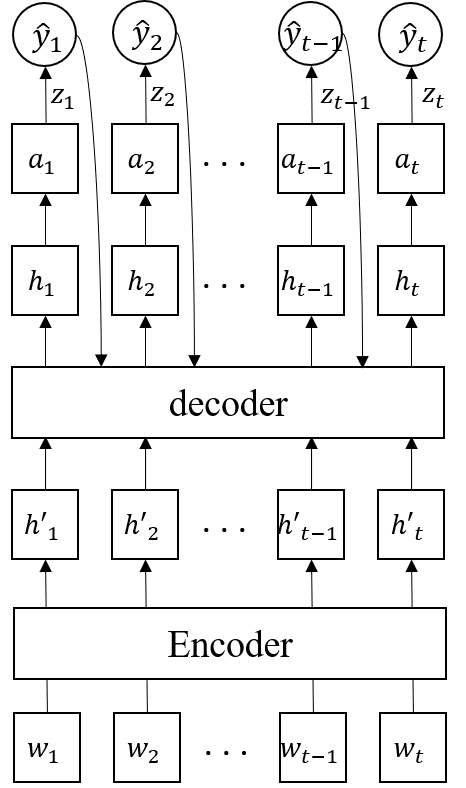}}
\caption{recurrent neural network (RNN)-based architecture for sequence labeling}
\end{figure}
\subsection{AL for sequence}
There are some existing active learning approaches that already show promising performance for sequence labeling. Among these approaches, we considered two uncertainty-based methods and one diversity-based method as baseline. Specifically, the uncertainty based methods we reviewed are: Maximum normalized log-probability (MNLP) and BALD, which exhibit a state-of-the-art performance \cite{ DBLP:conf/iclr/ShenYLKA18, DBLP:conf/emnlp/SiddhantL18}. In addition, the diversity-based method we considered herein adopts a coreset algorithm, instead of performing a direct evaluation of the classification task \cite{DBLP:conf/iclr/SenerS18}.\\
\textbf{Maximum normalized log-probability (MNLP)}: MNLP extends the least confidence (LC) method used for sequence selection using log probability normalized by sequence length and removing the bias observed in the LC model, which preferentially selects longer sentences \cite{DBLP:conf/iclr/ShenYLKA18}. Let \begin{math} n\end{math} be the length of the sequence of \begin{math} x_i\end{math}. Then, 
\begin{equation}
\frac{1}{n}\sum_{i=1}^{n}\max_{y_i, .. y_n}\log{P(y_i | y_1,.., y_n ,\{x_{i}\})}
\end{equation}
\textbf{Bayesian AL by disagreement (BALD)}: The Monte Carlo (MC) variant of BALD exploits an interpretation of dropout regularization as a Bayesian approximation to a Gaussian process \cite{DBLP:conf/icml/GalIG17, DBLP:conf/emnlp/SiddhantL18}. During inference, a fixed number of forward passes is executed with a dropout. The measure of uncertainty is the fraction of models across MC dropout samples from the network that disagree with the most popular choice. 
Let \begin{math}\widetilde{y}^{1}_j\end{math} represent the prediction of tags applied to the \begin{math} t\end{math}th forward pass on \begin{math} j\end{math}th sample, and \begin{math}T\end{math} be the number of forward passes. Then, 
\begin{equation}
\operatorname*{argmin}_j \left(1-\frac{\operatorname{count}(\operatorname{mode}(\widetilde{y}^{1}_j, ... , \widetilde{y}^{T}_j))}{T}\right)
\end{equation}
In this paper, we considered T = 100 independent dropout masks.

\textbf{Core-set}: Core-set is a pure diversity-based approach to find a subset of points, which is termed core-set, such that for all points, the maximum distance to the closest selected point is minimized using core-set selection \cite{DBLP:conf/iclr/SenerS18}. While the core-set is proposed for the convolutional neural network in the classification task, we adopted it for sequence labeling tasks with an RNN network. We leveraged the greedy furthest-first traversal condition on all labeled examples within the embedding; moreover, each example was computed by the encoder layer.

\section{Diverse Gradient Embedding for Sequences}

In this section, we introduce a method that simultaneously captures both uncertainty and diversity for active learning with sequence tagging. BADGE only addresses the image classification task. Moreover, it does not consider tasks that have sequences. In this section, first, we discuss how the gradient embedding of the last layer is related to the representation information. Second, we propose a simple modified sampling algorithm with length-normalized weight for improved sampling efficiency.
\begin{algorithm}[H]
\textbf{Input:} Unlabeled dataset U, sequence length of example L, number of budgets in query K. 
\begin{tabbing}
1. \=For all samples in U\\
\>1. Compute the hypothesis label \begin{math} \widehat{y} \end{math} \\
\>2. Compute the gradient embedding \begin{math} g_x \end{math} of the \\last layer
\end{tabbing}
2. Select data using weighted k-means++ seeding algorithm with L on \begin{math} g_x \end{math} until getting K number of samples.
 \caption{Active learning with diverse gradient embeddings for sequence labeling tasks.}
\end{algorithm}
Similar to \citet{DBLP:conf/iclr/AshZK0A20}, our method, described in Algorithm 1, performs two main computations at each AL round: (i) extracting gradient embeddings with respect to the parameters of final layers for all the unlabeled samples, and (ii) sampling a batch of query points based on these gradient embeddings using the weighted k-means++ initialization. It selects sentences for which the diversity and uncertainty are high when considering its length. 
The details of each computation to adapt it to the sequence labeling task is descried as follows:

\subsection{Gradient Embedding}

Here, we describe the gradient embedding of the penultimate layer in the sequence tagging task. Deep neural networks are optimized using gradient-based methods; therefore, training the gradient, back-propagated to a set of model parameters, captures the uncertainty of an example x, which is labeled with y. This may be viewed as a measure of change.

As the true label in the AL setting is unknown, \citet{DBLP:conf/aaai/ZhangLW17} investigated AL for sentence classification with expected gradient length over all possible classes. We computed the gradient assuming that the model’s current prediction on the example is its true label because its norm provides a lower bound on the gradient norm for the true label \cite{DBLP:conf/iclr/AshZK0A20}.

For given labels \begin{math}y \end{math} and sample sequence \begin{math}x\end{math}, this can minimize the negative log-likelihood of the objective function \begin{math}L_{NL} (x,y) \end{math} for the sequential labeling neural network. We denoted the nonlinear function that maps an input \begin{math}x\end{math} to output \begin{math}h_t \end{math} of the network’s penultimate layer that has a hidden state of current observation at time \begin{math}t\end{math} as \begin{math}z_t\end{math}.
% 여기 다시 써야함  not readable
In the last layer, \begin{math}a_t= W_{hz} h_t+b_z\end{math} where \begin{math}W_{hz}\end{math} is the weights of the layer and \begin{math}b_z\end{math} is bias. \begin{math}z_t=softmax(a_t) \end{math} is the nonlinearity softmax that predicts the probability assigned to the \begin{math}K\end{math} classes.
\begin{equation}
L_{NL} (x,y)=\sum_{t}y_t logz_t
\end{equation}
Note that the weight of the last layer, \begin{math}W_{hz}\end{math} is shared across all time sequences; consequently, we can differentiate \begin{math}W_{hz}\end{math} at each time step. 
If \begin{math}g_{x_t}^i\end{math} is defined as the derivative of the last layer weight, \begin{math} W_{hz}\end{math}, for \begin{math}L_{NL}(x,y)\end{math}, and the hypothesis label \begin{math}\widehat{y}_t = argmax(z_t)\end{math} is assumed, then the \begin{math}i\end{math}th block gradient corresponding to label \begin{math}i\end{math} is 
\begin{equation}
(g_x)_i = \frac{\partial }{\partial W_{hz}}L_{NL}(x,\widehat{y}) = \sum_{t}-(\widehat{y}_{ti}-p_{ti})h_t
\end{equation}

 Based on Equation 4, each block of \begin{math}g_x\end{math} is the sum of the product of the output of the penultimate layer and the probability vector at each time step. Although traditional representation learning in RNN uses only the final hidden state in the last step, the summation of each hidden state retains the property of representation information. In this respect, \begin{math}g_x\end{math} captures sample \begin{math}x\end{math}’s representation and has the probability scaling. It measures uncertainty as the gradient magnitude with respect to the parameters in the final layer.
\subsection{Diversity Based Sampling }
To capture diversity, BADGE uses k-means++ initialization sampling \cite{DBLP:conf/soda/ArthurV07} that favors both high magnitude and geometrical diversity based on gradient embeddings \begin{math}g_x\end{math}. In particular, it performs sequential sampling of k centers, where each new center is sampled from the ground set with probability proportional to the squared distance \begin{math}d(x)\end{math} to its nearest center. 
Moreover, each sentence has the same weight for classification tasks because we measured the budget in sentences. In sequence labeling tasks, however, we considered that the cost of annotating is proportional to the number of words because the annotator must provide one tag per word and that every word in the selected sentence must be annotated at once \cite{DBLP:conf/iclr/ShenYLKA18}. Therefore, it is important to select representative samples while considering the length of the sentence. Accordingly, we modified k-means++ initialization to weight the length of each sample.

\textbf{Weighted K-means++}: We modified the probability to sample the new center by weighted probability with the weight of the sentence length \begin{math}w(x)\end{math} of \begin{math}x\end{math} data. Denoting the length of \begin{math}x\end{math} data as \begin{math}l(x)\end{math}, \begin{math}w(x)\end{math} can be expressed as
\begin{math} w(x) = l(x)^2 / \sum_{x}l(x)^2\end{math}. In other words,
\begin{equation}
p(x)= \frac{w(x)d(x)^2}{\sum_{x}w(x)d(x)^2} .
\end{equation}
We abbreviated AL for diverse gradient embeddings with weighted k-means++ as W-BADGE.

\section{Experimental Setting}

\subsection{Datasets}
For the experiments, we used three publicly available datasets: CoNLL 2003 \cite{DBLP:conf/conll/SangM03} for the named entity recognition task, ATIS \cite{DBLP:conf/naacl/HemphillGD90} and Facebook Multilingual Task Oriented Dataset \cite{DBLP:conf/naacl/SchusterGSL19} dataset for the slot-filling task. We followed the standard splits dataset train/validation/test. CoNLL 2003 English NER data contains 14,041/3,250/3,453 samples for the trainvalidation/test sets. This dataset contains four different types of named entities: PERSON, LOCATION, ORGANIZATION, and MISC. ATIS comprises conversations from the airline domains. It contains 4,478/500/893 utterances for train/validation/test sets with 11 types of slots. The FMTOD dataset comprises multiple domains including alarm, reminder, and weather. We used the English data in FMTOD, which have 30,521/4,181/8,621 datasets and 11 types of slots. 
 \begin{figure*}[t]
 \centering
 \subfloat{\includegraphics[width=0.5\textwidth]{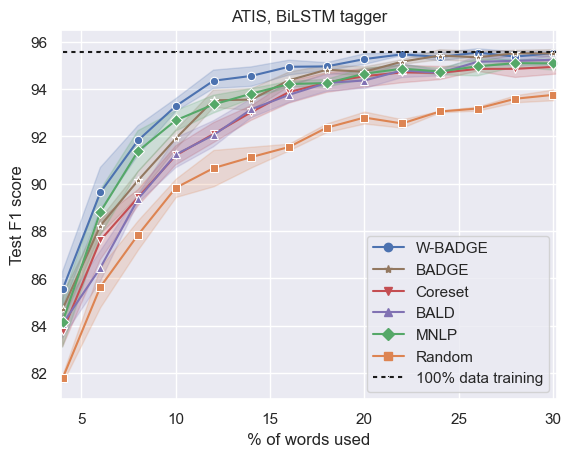}} 
 \subfloat{\includegraphics[width=0.5\textwidth]{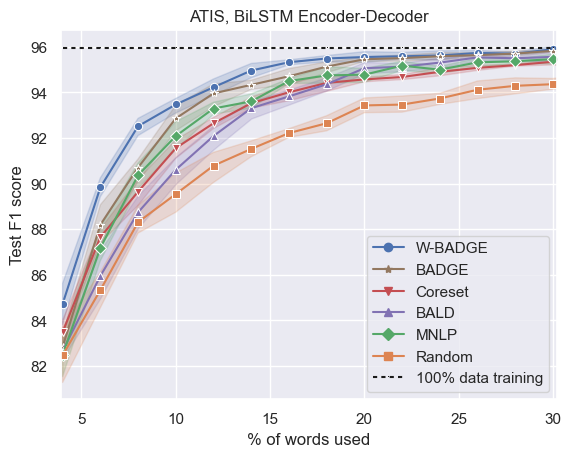}}
\newline
\noindent
 \subfloat{\includegraphics[width=0.5\textwidth]{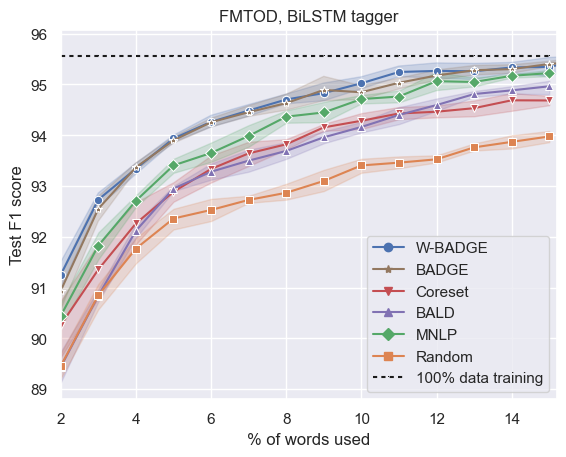}} 
 \subfloat{\includegraphics[width=0.5\textwidth]{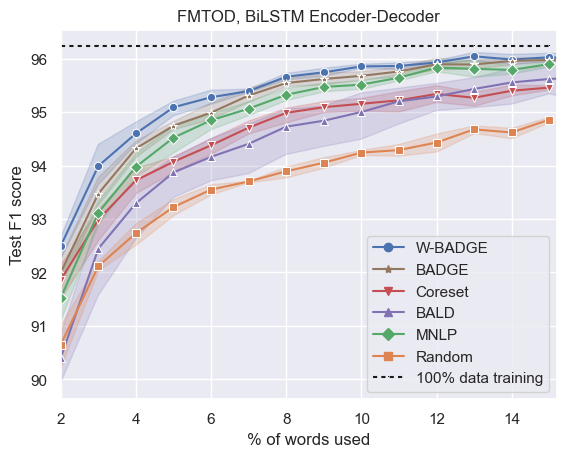}}
 \caption{Performance comparison of different datasets and various AL sampling methods used for the slot-filling task}
\end{figure*}

\begin{figure*}[t!]
 \centering
 \subfloat{\label{fig:}\includegraphics[width=0.5\textwidth]{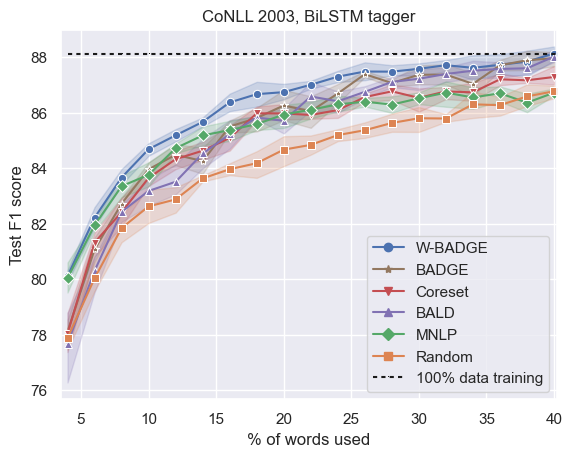}} 
 \subfloat{\label{fig:tiger}\includegraphics[width=0.5\textwidth]{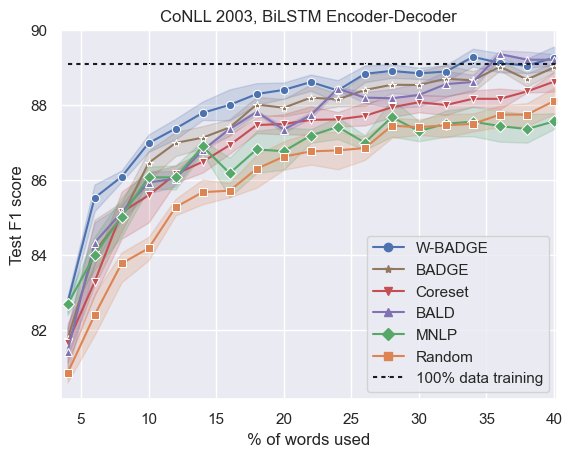}}
 \caption{Performance comparison of CoNLL2003 dataset and various AL sampling methods in the case of NER task}
\end{figure*}
\subsection{Model and Hyperparameter}
The two architectures used for training were the BiLSTM tagger (BiLSTM for word-level encoding) and BiLSTM encoder-decoder (BiLSTM for word-level encoding, and LSTM for decoding) \cite{DBLP:conf/icassp/ZhuY17} for both NER and slot-filling tasks, as shown in Figure 2. 
For the encoder, BiLSTM for word-level encoding contained 200 hidden units for ATIS and CoNLL, whereas it had 256 hidden units for FMTOD. Our model was comprised of 400-dimension word embeddings and utilized pre-trained GloVe and Kazuma \cite{DBLP:conf/emnlp/HashimotoXTS17}. Furthermore, we comprised a BiLSTM decoder with 100 dimensions for all datasets and utilized the greedy search for decoding. An Adam optimizer \cite{DBLP:journals/corr/KingmaB14} with a learning rate of 0.001 and a dropout rate of 0.5 were used to train our entire deep learning system. Additionally, we used 20, 16, and 32 batches for ATIS, CoNLL, and FMTOD, respectively. We trained model for 50 epochs for ATIS and FMTOD, whereas this number was 25 for CoNLL.

\subsection{Training Configuration}
 The AL process begins with initial samples randomly selected from the training dataset. We trained the initial model using this data. The learning process, which follows subsequently, consists of multiple rounds. At the beginning of each round, the AL algorithm selects sentences from the remaining training data to be annotated up to the predefined budget. After labeling the annotations, they are added to the training data. Accordingly, the model parameters were updated by training it on the new training dataset before proceeding to the next iteration. In each round, we trained the model from scratch to prevent overfitting \cite{DBLP:conf/iclr/HuLAR19}. We began our experiments with an initial labeled pool with 2\% labels of original training data for the
CoNLL and ATIS datasets, and 1\% labels of those for the FMTOD.
Further, we added the same number of labels at each iteration of active learning and evaluated the performance of the algorithm by its F1 score on the test dataset. We also reported the performance achieved after the full training of our model. All experiments were repeated five times and the average F-scores with their standard deviations are reported.

\section{Results}
We evaluated the performances of various AL methods with different models for the slot-filling and NER tasks and plotted them in Figures 2 and 3, respectively. The x-axis represents the percent of words annotated and used for training and the y-axis indicates the best F1 scores obtained. Moreover, standard errors are indicated by shaded regions. In all cases, we observed that the W-BADGE method indicated a significant improvement or over the baseline AL methods. It consistently outperforms those employing either pure uncertainty or diversity-based sampling in both the earlier and latter rounds. The performance gap is clearer, especially in the earlier rounds of training. Although BADGE is comparable with other methods, W-BADGE performs as well as or better than BADGE.
\\
More specifically, as can be observed from Figure 2, W-BADGE performs satisfactorily in both the ATIS and FMTOD datasets and shows significant improvement in all baselines when performing the slot-filling task. While BADGE also shows the similar performance in FMTOD dataset with BiLSTM tagger, W-BADGE slightly outperformed in BiLSTM Encoder-Decoder model. Moreover, classic uncertainty-based sampling MNLP outperforms BALD and Coreset in both datasets. We suppose that the slot-filling task is relatively complex and sparse because it has more classes than the NER task; therefore, uncertainty-based sampling methods have an advantage when performing such a task. 
\\
Figure 3 graphs the learning curves of the AL algorithms on CoNLL 2003 dataset for NER task. We observed that W-BADGE presented significant improvement over BALD, which shows the best performance in NER tasks \cite{DBLP:conf/emnlp/SiddhantL18} and BADGE. As W-BADGE outperformed the Coreset, it appears that W-BADGE can be advantageous not only over purely diversity-based approaches but also over classic uncertainty-based approaches in NER task. 
\\
Overall, the advantages of weighted diverse gradient embeddings can be substantial. For example, We find that active learning algorithms achieve 98-99\% deep model trained on full data using only 15\% of samples in ATIS dataset. The relative improvement remains significant over baselines.

\section{Conclusion}
In this study, we explored the empirical study on AL utilizing the advantages of both uncertainty and diversity by selecting weighted diverse gradient embeddings to perform a sequence labeling task. We proposed an efficient method and empirically demonstrated that it could consistently achieve a superior performance while consuming much less data. It adds robustness to the dataset and the architecture, thus proving to be a useful option for solving real-world active learning problems

\bibliography{anthology,aacl-ijcnlp2020}
\bibliographystyle{acl_natbib}

\appendix

\end{document}